\documentclass[10pt,twocolumn,letterpaper]{article}

\usepackage{iccv}
\usepackage{times}
\usepackage{epsfig}
\usepackage{graphicx}
\usepackage{amsmath}
\usepackage{amssymb}
\usepackage{stfloats}

\usepackage[accsupp]{axessibility}

\usepackage[breaklinks=true,bookmarks=false]{hyperref}

\iccvfinalcopy 

\def\iccvPaperID{****} 

\ificcvfinal\fi

\begin{document}

\title{TPH-YOLOv5: Improved YOLOv5 Based on Transformer
Prediction Head for Object Detection on Drone-captured Scenarios}

\author{
Xingkui Zhu\textsuperscript{1} \textsuperscript{*} \and Shuchang Lyu\textsuperscript{1} \thanks{Contribute Equally.} \and Xu Wang \textsuperscript{1} \and Qi Zhao\textsuperscript{1} \thanks{Corresponding author.} 
\and \\
\textsuperscript{1} Beihang University, Beijing, China \\
{\tt\small \{adlith, lyushuchang, sy2002406, zhaoqi\}@buaa.edu.cn}
}

\maketitle
\ificcvfinal\thispagestyle{empty}\fi

\begin{abstract}
Object detection on drone-captured scenarios is a recent popular task. As drones always navigate in different altitudes, the object scale varies violently, which burdens the optimization of networks. Moreover, high-speed and low-altitude flight bring in the motion blur on the densely packed objects, which leads to great challenge of object distinction. To solve the two issues mentioned above, we propose TPH-YOLOv5. Based on YOLOv5, we add one more prediction head to detect different-scale objects. Then we replace the original prediction heads with Transformer Prediction Heads (TPH) to explore the prediction potential with self-attention mechanism. We also integrate convolutional block attention model (CBAM) to find attention region on scenarios with dense objects. To achieve more improvement of our proposed TPH-YOLOv5, we provide bags of useful strategies such as data augmentation, multi-scale testing, multi-model integration and utilizing extra classifier. Extensive experiments on dataset VisDrone2021 show that TPH-YOLOv5 have 
good performance with impressive interpretability on drone-captured scenarios. On DET-test-challenge dataset, the AP result of TPH-YOLOv5 are 39.18\%, which is better than previous SOTA method (DPNetV3) by 1.81\%. On VisDrone Challenge 2021, TPH-YOLOv5 wins $5^{th}$ place and achieves well-matched results with $1^{st}$ place model (AP 39.43\%). Compared to  baseline model (YOLOv5), TPH-YOLOv5 improves about 7\%, which is encouraging and competitive.
\end{abstract}

\begin{figure}[t]
\begin{center}
   \includegraphics[width=0.8\linewidth]{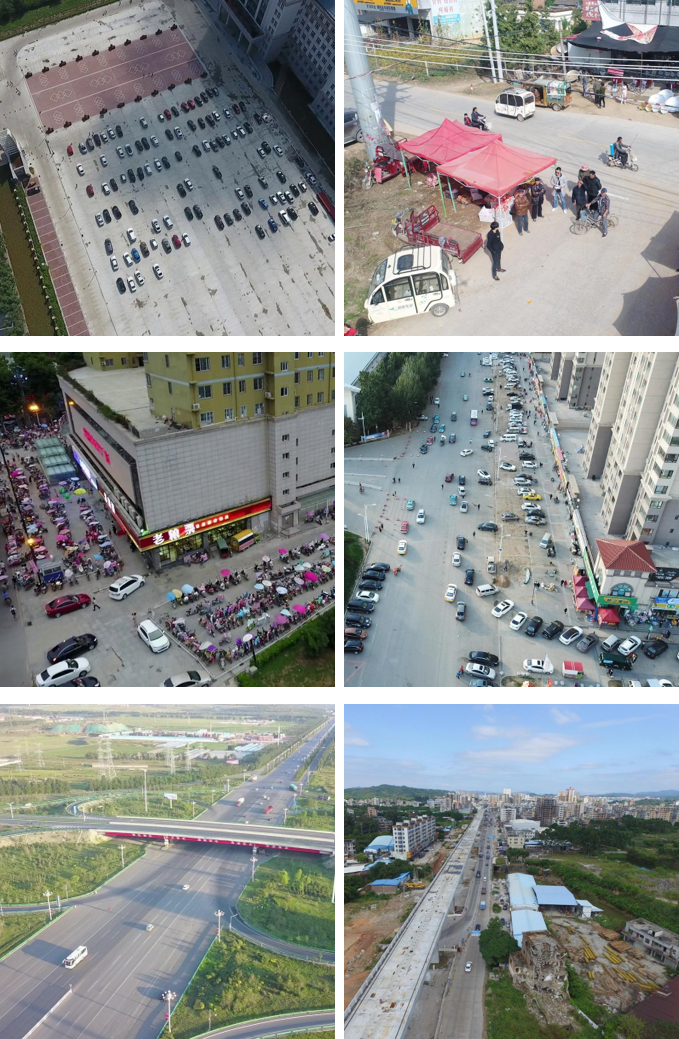}
\end{center}
   \caption{Intuitive cases to explain the three main problems in object detection on drone-captured images. The cases in first row, second row and third row respectively shows the size variation, high-density and large coverage of objects on drone-captured images.}
\label{Fig1}
\end{figure}

\section{Introduction}

Object detection technology on drone-captured scenarios has been widely used in many practical applications, such as plant protection~\cite{UAV_detect1, UAV_detect2}, wildlife protection~\cite{UAV_detect3, UAV_detect4} and urban surveillance~\cite{UAV_detect5, UAV_detect6}. In this paper, we focus on improving the performance of object detection on drone-captured images and providing insight for the above-mentioned numerous applications.
\par Recent years have witnessed significant progresses in object detection tasks using deep convolutional neural networks~\cite{ren2015faster, redmon2016you, liu2016ssd, RetinaNet, VFNet}. Some notable benchmark datasets like MS COCO~\cite{lin2014microsoft} and PASCALVOC~\cite{VOC} greatly promote the development of object detection application.
However, most previous deep convolutional neural networks are designed for natural scene images. Directly applying previous models to tackle object detection task on drone-captured scenarios mainly has three problems, which are intuitively illustrated by some cases in Fig.\ref{Fig1}. First, the object scale varies violently because the flight altitude of drones change greatly. Second, drone-captured images contain objects with high density, which brings in occlusion between objects. Third, drone-captured images always contain confusing geographic elements because of covering large area. The above-mentioned three problems make the object detection of drone-captured images very challenging.

\begin{figure*}[ht]
\begin{center}
   \includegraphics[width=0.8\linewidth]{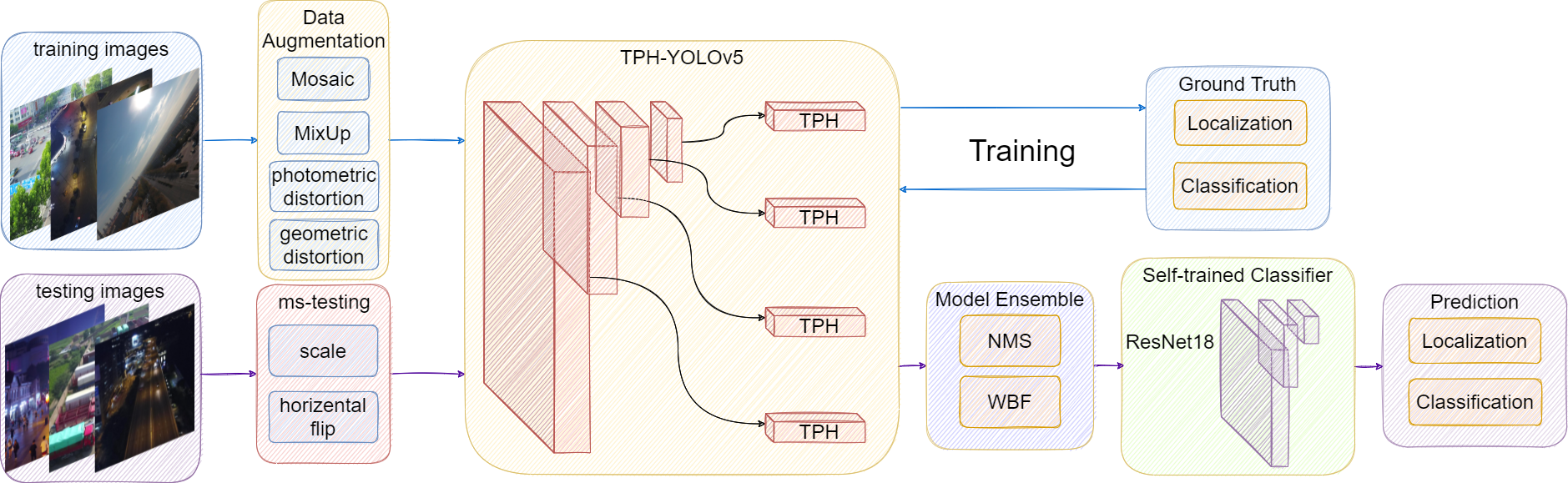}
\end{center}
   \caption{The overview of working pipeline using TPH-YOLOv5. Compared to original version, we mainly improve the head by applying Transformer Prediction Head (TPH). We also add one more head to better detect different scale objects. In addition, we employ bag of tricks like data augmentation, multi-scale testing, model ensemble and self-trained classifier to make TPH-YOLOv5 stronger.}
\label{Fig2}
\end{figure*}

\par In object detection task, YOLO series~\cite{redmon2016you,redmon2017yolo9000,redmon2018yolov3,bochkovskiy2020yolov4} play an important role in one-stage detectors. In this paper, we propose an improved model, TPH-YOLOv5 based on YOLOv5~\cite{glenn_jocher_2021_4679653} to solve the above-mentioned three problems. The overview of the detection pipeline using TPH-YOLOv5 is shown in Fig.\ref{Fig2}. We respectively use CSPDarknet53~\cite{wang2020cspnet,bochkovskiy2020yolov4} and path aggregation network (PANet~\cite{liu2018path}) as backbone and neck of TPH-YOLOv5, which follows the original version. In the head part, we first introduce one more head for tiny object detection. Totally, TPH-YOLOv5 contains four detection heads separately used for the detection of tiny, small, medium, large objects. Then, we replace the original prediction heads with Transformer Prediction Heads (TPH) ~\cite{vit, Transformer} to explore the prediction potential. To find the attention region in images with large coverage, we adopt Convolutional Block Attention Module (CBAM~\cite{woo2018cbam}) to sequentially generate the attention map along channel-wise and spatial-wise dimensions. Compared to YOLOv5, our improved TPH-YOLOv5 can better deal with drone-captured images.
\par To further improve the performance of TPH-YOLOv5, we employ bag of tricks (Fig.\ref{Fig2}). Specifically, we adopt data augmentation during training, which promote the adaptation for dramatic size changes of objects in images. We also add multi-scale testing (ms-testing) and multi-model ensemble strategies during inference to obtain more convincing detection results. Moreover, through visualizing the failure cases, we find that our proposed architecture has excellent localization ability but poor classification ability, especially on some similar categories like ``tricycle'' and ``awning-tricycle''. To solve this problem, we provide a self-trained classifier (ResNet18~\cite{he2016deep}) using the image patches cropping from training data as classification training set. With self-trained classifier, our method has 0.8\%$\sim$1.0\% improvement on AP value.
\par Our contributions are listed as follows:
\begin{itemize}
    \item We add one more prediction head to deal with large scale variance of objects.
    \item We integrate the Transformer Prediction Heads (TPH) into YOLOv5, which can accurately localize objects in high-density scenes.
    \item We integrate CBAM into YOLOv5, which can help the network to find region of interest in images that have large region coverage. 
    \item We provide useful bag of tricks and filtering some useless tricks for object detection task on drone-captured scenarios.
    \item We use self-trained classifier to improve the classification ability on some confusing categories.
    \item On VisDrone2021 test-challenge dataset, our proposed TPH-YOLOv5 achieve 39.18\% (AP), outperforming DPNetV3 (previous SOTA method) by 1.81\%. In VisDrone2021 DET challenge, TPH-YOLOv5 wins $5^{th}$ place and has minor gap comparing with $1^{st}$ place models.
\end{itemize}

\section{Related Work }

\subsection{Data Augmentation}
The effectiveness of data augmentation is to expand the dataset, so that the model has higher robustness to the images obtained from different environments.
Photometric distortions and geometric distortions are wildly used by researchers. As for photometric distortion, we adjusted the hue, saturation and value of the images. In dealing with geometric distortion, we add random scaling, cropping, translation, shearing, and rotating.
In addition to the above-mentioned global pixel augmentation methods, there are some more unique data augmentation methods. Some researchers have proposed methods using multiple images together for data augmentation \ie MixUp~\cite{zhang2017MixUp}, CutMix~\cite{yun2019cutmix} and Mosaic~\cite{bochkovskiy2020yolov4}.
MixUp randomly select two samples from the training images to perform random weighted summation, and the labels of the samples also correspond to the weighted summation.
Unlike occlusion works that generally use zero-pixel "black cloth" to occlude a image, CutMix uses an area of another image to cover the occluded area.
Mosaic is an improved version of the CutMix. Mosaic stitches four images, which greatly enriches the background of the detected object. In addition, batch normalization calculates the activation statistics of 4 different images on each layer.

In TPH-YOLOv5, we use a combination of MixUp, Mosaic and traditional methods in data augmentation.

\subsection{Multi-Model Ensemble Method in Object Detection}
Deep learning neural networks are non-linear methods. They provide greater flexibility and can scale in proportion to the amount of training data. One disadvantage of this flexibility is that they learn through random training algorithms, which means that they are sensitive to the details of the training data, and may find a different set of weights each time they train, resulting in different predictions. This gives the neural network a high variance. A successful way to reduce the variance of neural network models is to train multiple models instead of a single model, and combine the predictions of these models.

There are three different methods to ensemble boxes from different object detection models: Non-maximum suppression (NMS)~\cite{neubeck2006efficient}, Soft-NMS~\cite{wang2021daedalus}, weighted boxes fusion (WBF)~\cite{solovyev2021weighted}. 
In the NMS method, if the overlap, intersection over union (IoU) of the boxes is higher than a certain threshold, they are considered to belong to the same object. For each object, NMS only leaves one bounding box with the highest confidence, and other bounding boxes are deleted. Therefore, the box filtering process depends on the choice of this single IoU threshold, which have a big impact on model performance.
Soft-NMS has made a slightly change to NMS, which made Soft-NMS shows a significant improvement over traditional NMS on standard benchmark datasets (such as PASCAL VOC~\cite{everingham2010pascal} and MS COCO~\cite{lin2014microsoft}). It sets an attenuation function for the confidence of adjacent bounding boxes based on the IoU value instead of completely setting their confidence scores to zero and delete them.
WBF works differently from NMS. Both NMS and Soft-NMS exclude some boxes, while WBF merges all boxes to form the final result. Therefore, it can solve all the inaccurate predictions of the model.
We use WBF to ensemble final models, which performs much better than NMS.

\subsection{Object Detection}
CNN-based object detectors can be divided into many types: 1) one-stage detectors: YOLOX~\cite{ge2021yolox}, FCOS~\cite{tian2019fcos}, DETR~\cite{zhu2020deformable}, Scaled-YOLOv4~\cite{wang2021scaled}, EfficientDet~\cite{tan2020efficientdet}. 
2) two-stage detectors: VFNet~\cite{zhang2021varifocalnet}, CenterNet2~\cite{zhou2021probabilistic}.
3) anchor-based detectors: Scaled-YOLOv4~\cite{wang2021scaled}, YOLOv5~\cite{glenn_jocher_2021_4679653}.
4) anchor-free detectors: CenterNet~\cite{zhou2019objects}, YOLOX~\cite{ge2021yolox}, RepPoints~\cite{yang2019reppoints}. 
Some detectors are specially designed for Drone-captured images like RRNet~\cite{chen2019rrnet}, PENet~\cite{tang2020penet}, CenterNet~\cite{zhou2019objects} \etc.
But from the perspective of components, they generally consist of two parts, an CNN-based backbone, used for image feature extraction, and the other part is detection head used to predict the class and bounding box for object.
In addition, the object detectors developed in recent years often insert some layers between the backbone and the head, people usually call this part the neck of the detector.
Next, we will separately introduce these three structures in detail.

\noindent \textbf{Backbone.}
The backbone that are often used include VGG~\cite{simonyan2014very}, ResNet~\cite{he2016deep}, DenseNet~\cite{huang2017densely}, MobileNet~\cite{howard2017mobilenets}, EfficientNet~\cite{tan2019efficientnet}, CSPDarknet53~\cite{wang2020cspnet}, Swin Transformer~\cite{liu2021swin} \etc , rather than networks designed by ourselves. Because these networks have proven that they have strong feature extraction capabilities on classification and other issues. But researchers will also fine-tune the backbone to make it more suitable for specific tasks.

\noindent \textbf{Neck.}
The neck is designed to make better use of the features extracted by the backbone. It reprocesses and rationally uses the feature maps extracted by Backbone at different stages. Usually, a neck consists of several bottom-up paths and several top-down paths. Neck is a key link in the target detection framework.
The earliest neck is the use of up and down sampling block. The feature of this method is that there is no feature layer aggregation operation, such as SSD~\cite{liu2016ssd}, directly follow the head after the multi-level feature map.
Commonly used path-aggregation blocks in neck are: FPN~\cite{lin2017feature}, PANet~\cite{liu2018path}, NAS-FPN~\cite{ghiasi2019fpn}, BiFPN~\cite{tan2020efficientdet}, ASFF~\cite{liu2019learning}, SFAM~\cite{zhao2019m2det}.The commonality of these methods is to repeatedly use various up-and-down sampling, splicing, dot sum or dot product to design aggregation strategies.
There are also some additional blocks used in neck, like SPP~\cite{he2015spatial}, ASPP~\cite{chen2017deeplab}, RFB~\cite{liu2018receptive}, CBAM~\cite{woo2018cbam}.

\noindent \textbf{Head.}
As a classification network, the backbone cannot complete the positioning task, and the head is designed to be responsible for detecting the location and category of the object by the features maps extracted from the backbone. Heads are generally divided into two kinds: one-stage object detector and two-stage object detector. 
Two-stage detectors have long been the dominant method in the field of object detection, and the most representative one is the R-CNN series~\cite{girshick2014rich,girshick2015fast,ren2015faster}. Compared with the two-stage detector, the one-stage detector predicts the bounding box and the class of objects at the same time. The speed advantage of the one-stage detector is obvious, but the accuracy is lower. For one-stage detectors, the most representative models are YOLO series~\cite{redmon2016you,redmon2017yolo9000,redmon2018yolov3,bochkovskiy2020yolov4}, SSD~\cite{liu2016ssd} and RetinaNet~\cite{lin2017focal}.

\section{TPH-YOLOv5}

\begin{figure*}[htp]
\begin{center}
\includegraphics[width=0.95\linewidth]{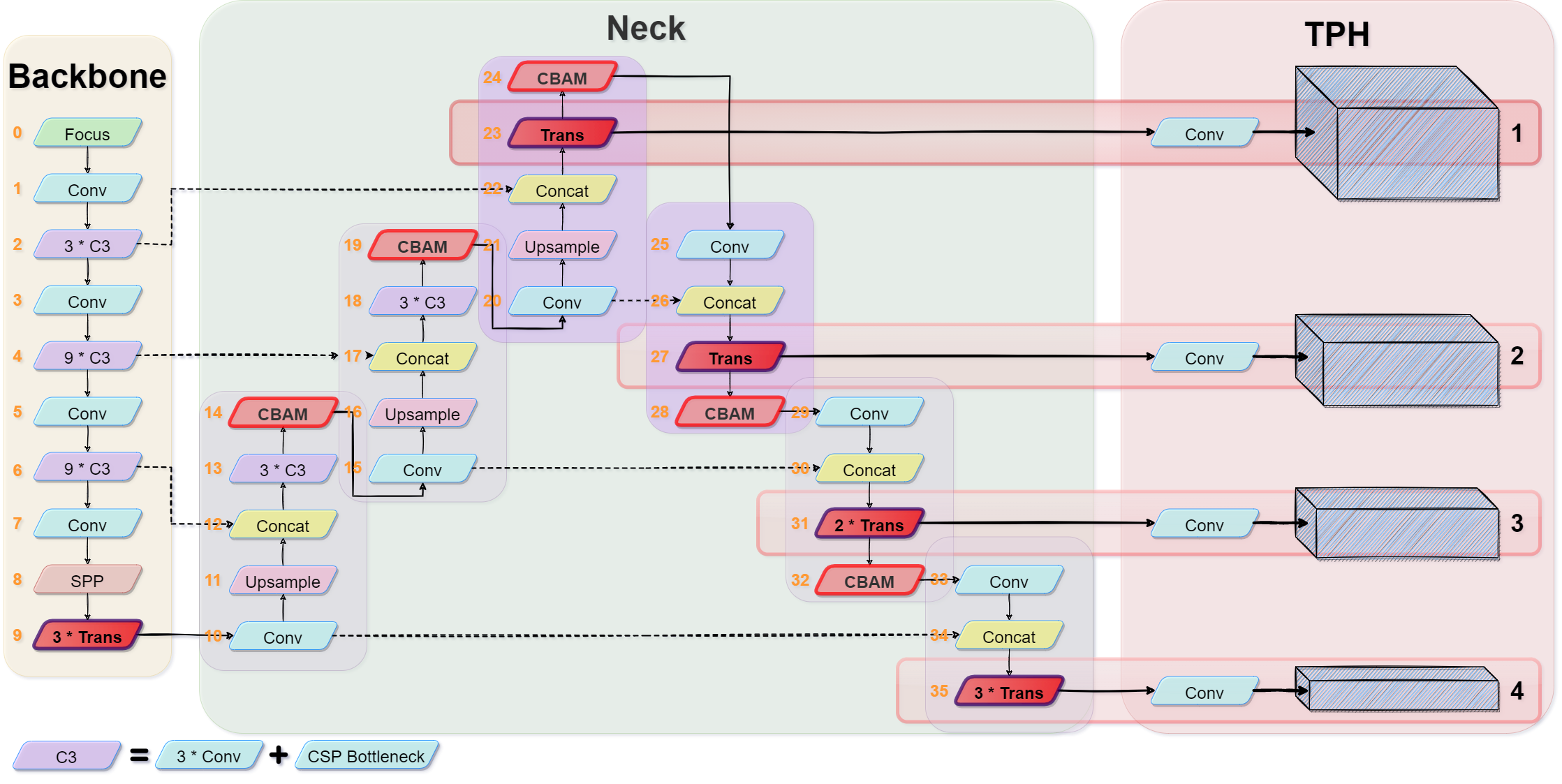}
\end{center}
   \caption{The architecture of the TPH-YOLOv5. a) CSPDarknet53 backbone with three transformer encoder blocks at the end. b) The Neck use the structure like PANet. c) Four TPHs (transformer prediction heads) use the feature maps from transformer encoder blocks in Neck. In addition, the number of each block is marked with orange numbers on the left side of the block.}
\label{fig:tph-yolov5}
\end{figure*}
\subsection{Overview of YOLOv5}
 \par YOLOv5 has four different models including YOLOv5s, YOLOv5m, YOLOv5l and YOLOv5x. Generally, YOLOv5 respectively uses the architecture of CSPDarknet53 with an SPP layer as backbone, PANet as Neck and YOLO detection head~\cite{redmon2016you}. To further optimize the whole architecture, bag of freebies and specials~\cite{bochkovskiy2020yolov4} are provided. Since it is the most notable and convenient one-stage detector, we select it as our baseline.
 \par When we train the model using VisDrone2021 dataset~\cite{zhu2018vision} with data augmentation strategy (Mosaic and MixUp), we find that the results of YOLOv5x are much better than YOLOv5s, YOLOv5m and YOLOv5l, and the gap of AP value is more than 1.5\%. Even though the training computation cost of the YOLOv5x model is more than that of other three models, we still choose to use YOLOv5x to pursue the best detection performance. In addition, according to the features of drone-captured images, we adjust the parameters of commonly used photometric distortions and geometric distortions.

\subsection{TPH-YOLOv5}
\par The framework of TPH-YOLOv5 is illustrated in Fig.~\ref{fig:tph-yolov5}. We modify the original YOLOv5 to make it specialize in the VisDrone2021 dataset.
\par \noindent \textbf{Prediction head for tiny objects.}
We investigate the VisDrone2021 dataset and find that it contains many extremely small instances, so we add one more prediction head for tiny objects detection. Combined with the other three prediction heads, our four-head structure can ease the negative influence caused by violent object scale variance. As shown in Fig.~\ref{fig:tph-yolov5}, the prediction head (head No.1) we add is generated from low-level, high-resolution feature map, which is more sensitive to tiny objects. After adding an additional detection head, although the computation and memory cost increase, the performance of tiny objects detection gets large improvement.

\begin{figure}[htbp]
\begin{center}
\includegraphics[width=0.3\linewidth]{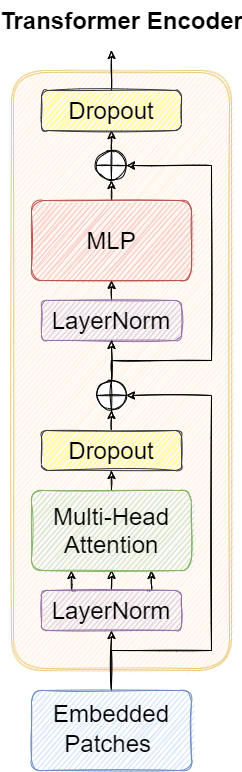}
\end{center}
   \caption{The architecture of transformer encoder, which contains two main blocks, a multi-head attention block and a feed-forward neural network (MLP). LayerNorm and Dropout layers help the network converge better and prevent the network from over fitting. Multi-head attention can help the current node not only pay attention to the current pixels, but also obtain the semantics of the context.}
\label{fig:transformer-encoder}
\end{figure}

\noindent \textbf{Transformer encoder block.} Inspired by the vision transformer~\cite{dosovitskiy2020image}, we replace some convolutional blocks and CSP bottleneck blocks in original version of YOLOv5 with transformer encoder blocks. The structure is shown in Fig.~\ref{fig:transformer-encoder}. Compared to original bottleneck block in CSPDarknet53, we believe that transformer encoder block can capture global information and abundant contextual information. Each transformer encoder contains two sub-layers. The first sub-layer is a multi-head attention layer and the second one (MLP) is a fully-connected layer. Residual connections are used between each sub-layer. Transformer encoder blocks increase the ability to capture different local information. It can also explore the feature representation potential with self-attention mechanism~\cite{vaswani2017attention}. On the VisDrone2021 dataset, transformer encoder blocks have better performance on occluded objects with high-density.
\par Based on YOLOv5, we only apply transformer encoder blocks in the head part to form Transformer Prediction Head (TPH) and the end of backbone. Because the feature maps at the end of the network have low resolution. Applying TPH on low-resolution feature maps can decrease the expensive computation and memory cost. Moreover, when we enlarge the resolution of input images, we optional remove some TPH blocks at early layers to make the training process available.

\begin{figure}[tbp]
\begin{center}
\includegraphics[width=1\linewidth]{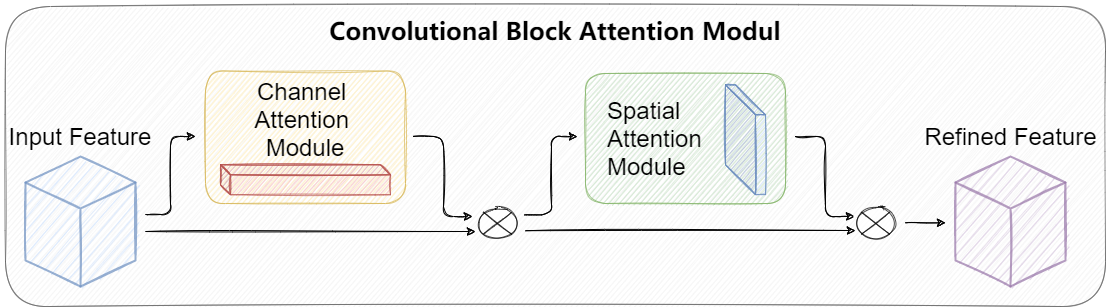}
\end{center}
   \caption{The overview of CBAM module. Two sequential sub-modules are used to refine feature map that go through CBAM, residual paths are also used.}
\label{fig:CBAM}
\end{figure}

\noindent \textbf{Convolutional block attention module (CBAM).}
CBAM~\cite{woo2018cbam} is a simple but effective attention module. It is a lightweight module that can be integrated into most notable CNN architectures, and it can be trained in an end-to-end manner. Given a feature map, CBAM sequentially infers the attention map along two separate dimensions of channel and spatial, and then multiplies the attention map with the input feature map to perform adaptive feature refinement. The structure of the CBAM module is shown in the Fig.~\ref{fig:CBAM}. According to the experiment in the paper~\cite{woo2018cbam}, after integrating CBAM into different models on different classification and detection datasets, the performance of the model get large improved, which proves the effectiveness of this module.
\par On drone-captured images, large covering region always contains confusing geographical elements. Using CBAM can extract the attention area to help TPH-YOLOv5 resist the confusing information and focus on useful target objects.

\noindent \textbf{Ms-testing and model ensemble.}
We train five different models in terms of different perspectives for model ensemble. During inference phase, we first perform ms-testing strategy on single model. The implementation details of ms-testing are the following three steps. 1) Scaling the testing image to 1.3 times. 2) Respectively reducing the image to 1 time, 0.83 times, and 0.67 times. 3) Flipping the images horizontally. Finally, we feed the six different-scaling images to TPH-YOLOv5 and use NMS to fuse the testing predictions. 
\par On different models, we perform the same ms-testing operation and fuse the final five predictions by WBF to get the final result.

\begin{figure}[htbp]
\begin{center}
\includegraphics[width=1\linewidth]{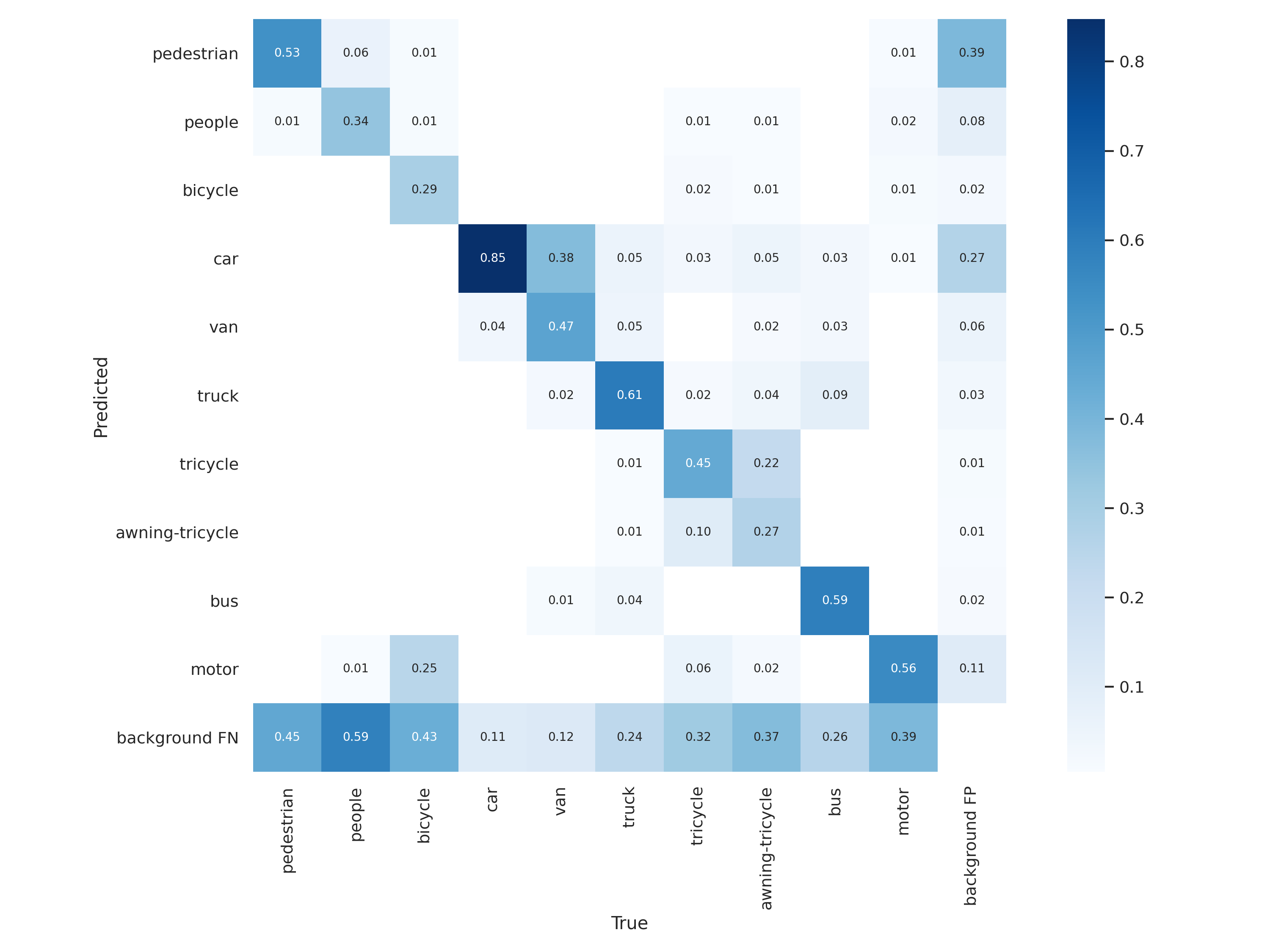}
\end{center}
   \caption{Confusion matrix was made at IoU threshold of 0.45, confidence threshold of 0.25.}
\label{fig:confusion matrix}
\end{figure}

\noindent \textbf{Self-trained classifier.}
After training the VisDrone2021 dataset with TPH-YOLOv5, we test the test-dev dataset and then analyze the results by visualizing the failure cases and draw a conclusion that TPH-YOLOv5 has excellent localization ability but poor classification ability. We further explore the confusion matrix which is shown in Fig.\ref{fig:confusion matrix}, and observe that the precision of the some hard categories such as tricycle and awning-tricycle are very low. Therefore, we propose an extra self-trained classifier. First, we construct a training set by cropping the ground-truth bounding boxes and resizing each image patches to 64$\times$64. Then we select ResNet18~\cite{he2016deep} as classifier network. As shown in experimental results, our method get around 0.8\%\textasciitilde1.0\% improvement on AP value with the help of this self-trained classifier.

\section{Experiments}
We use the testset-challenge and testset-dev of the VisDrone2021 dataset to evaluate our model, and we report mAP (average of all 10 IoU thresholds, ranging from [0.5: 0.95]) and AP50. VisDrone2021-DET dataset is the same as VisDrone2019-DET dataset and VisDrone2018-DET dataset.

\subsection{ Implementation Details}
We implement TPH-YOLOv5 on Pytorch 1.8.1. All of our models use an NVIDIA RTX3090 GPU for training and testing.
In the training phase, we use part of pre-trained model from yolov5x, because TPH-YOLOv5 and YOLOv5 share most part of backbone (block 0\textasciitilde8) and some part of head (block 10\textasciitilde13 and block 15\textasciitilde18), there are many weights can be transferred from YOLOv5x to TPH-YOLOv5, by using these weights we can save a lot of training time.

Because the VisDrone2021 training set is a bit small, we only train the model on VisDrone2021 trainset for 65 epochs, and the first 2 epochs are used for warm-up. We use adam optimizer for training, and use 3e-4 as the initial learning rate with the cosine lr schedule. The learning rate of the last epoch decays to 0.12 of the initial learning rate.
The size of the input image of our model is very large, the long side of the image is 1536 pixels, which leads to the batch size is only 2.

\begin{figure}[htbp]
\begin{center}
\includegraphics[width=1\linewidth]{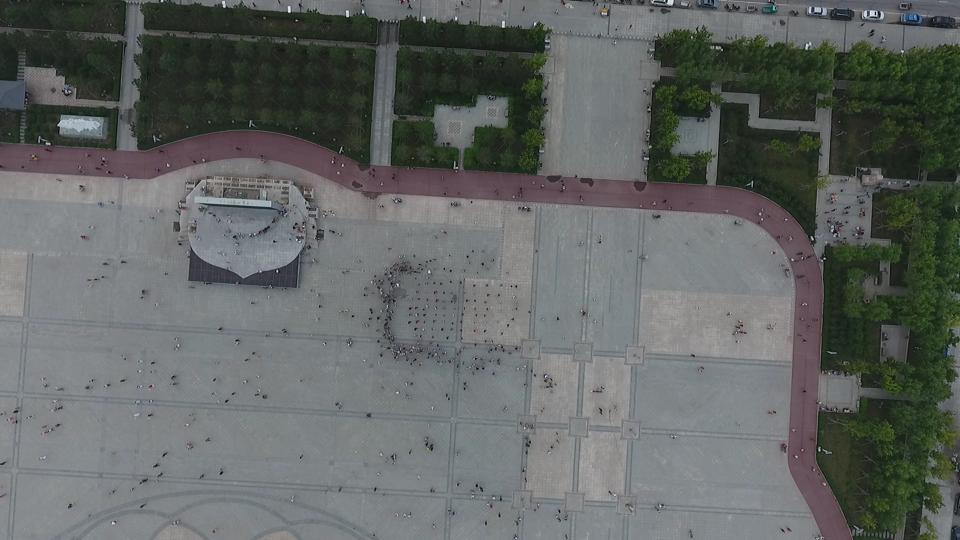}
\end{center}
   \caption{Some images were taken too high, resulting in many small objects, which cannot be recognized.}
\label{fig:exsmall}
\end{figure}

\noindent \textbf{Data analysis.}
According to our previous engineering experience, it is very important to walk through dataset before training the model, which can often be of great help to the improvement of mAP.
We have analyzed bounding boxes in the VisDrone2021 dataset. When the input image size is set to 1536, there are 622 of 342391 labels are less than 3 pixels in size. As shown in Fig.~\ref{fig:exsmall}, these small objects are hard to recognize. When we use gray squares to cover these small objects and train our model on the processed dataset, the mAP improves by 0.2, better than not.

\noindent \textbf{Ms-testing.}
When training neural network models for computer vision problems, data augmentation is a technique often used to improve performance and reduce generalization errors.
When using a model to make predictions, image data augmentation of test dataset can also be applied to allow the model to make predictions on multiple different versions of images. The prediction of the augmented images can be averaged to get better prediction performance.

We scale the test images to three different sizes in ms-testing, and then flip them horizontally, so that a total of 6 different images are obtained. After testing six different images and fusing the results, we get the final test result.


\subsection{Comparisons with the State-of-the-art}
\noindent \textbf{On VisDrone2021-DET testset-challenge.}

\begin{table}[h]
\begin{center}
\resizebox{0.45\textwidth}{!}{ 
\begin{tabular}{|l|c|c|}
\hline
Methods & mAP (\%) & AP50 (\%) \\
\hline\hline
RetinaNet\cite{lin2017focal} & 11.81 & 21.37 \\
RefineDet\cite{zhang2018single} & 14.90 & 28.76 \\
DetNet59\cite{li2018detnet}  & 15.26 & 29.23 \\
Cascade-RCNN\cite{cai2018cascade}  &  16.09 & 31.91 \\
FPN\cite{lin2017feature}  &  16.51 & 32.20 \\
Light-RCNN\cite{li2017light}  &  16.53 & 32.78 \\
CornetNet\cite{law2018cornernet}  &  17.41 & 34.12 \\
\hline
RRNet (2019 $2^{nd}$)\cite{chen2019rrnet}  &  29.13 & 55.82 \\
DPNet-ensemble (2019 SOTA)~\cite{du2019visdrone} & 29.62 & 54.00\\
SMPNet (2020 $2^{nd}$)\cite{2020leaderboard} & 35.98 &59.53 \\
DPNetV3 (2020 SOTA)\cite{2020leaderboard} & 37.37 & 62.05\\
TPH-YOLOv5 ensemble & \textbf{39.18} & $\backslash$ \\
\hline
\end{tabular}
}
\end{center}
\caption{The comparison of the performance in VisDrone2021 testset-challenge}
\label{table:testset-cahllenge}
\end{table}

Due to the limited number of submissions in the VisDrone2021 competition server, we only obtained the results of 4 models on testset-challenge and the final results of the ensemble of 5 models. 
We finally got a good score of 39.18 on testset-challenge, which is much higher than VisDrone2020's best score of 37.37.
Ranked fifth in the VisDrone 2021 leader board, our score is 0.25 lower than the 39.43 of the first place. If the number of submissions is not used up, we will definitely get better results.
Table \ref{table:testset-cahllenge} lists the score of our model, compared with the scores in the previous year’s VisDrone competition and the scores of algorithms submitted by the committee.

\subsection{Ablation Studies}

\noindent \textbf{On VisDrone2021-DET testset-dev.}
we analyze importance of each proposed component on local testset-dev as we cannot test these on VisDrone2021 competition server, the number of submissions to the competition server is very valuable. The impact of each component is listed in the table~\ref{table:testset-dev}.

\begin{table}[h]
\begin{center}
\resizebox{0.45\textwidth}{!}{ 
\begin{tabular}{|l|c|c|}
\hline
Methods & mAP (\%) & AP50 (\%)\\
\hline\hline
YOLOv5 & 28.88 & 49.33 \\
YOLOv5+P2 & 31.03 (\textcolor{green}{$\uparrow$2.15}) & 51.61 (\textcolor{green}{$\uparrow$ 2.28}) \\
YOLOv5+P2+transformer & 32.84 (\textcolor{green}{$\uparrow$ 1.81})&  53.87 (\textcolor{green}{$\uparrow$ 2.26}) \\
TPH-YOLOv5 (previous+CBAM) & 33.63 (\textcolor{green}{$\uparrow$ 0.79})& 54.77 (\textcolor{green}{$\uparrow$ 0.90}) \\
TPH-YOLOv5+ms-testing & 34.90 (\textcolor{green}{$\uparrow$ 1.27})& 56.40 (\textcolor{green}{$\uparrow$ 1.63}) \\
TPH-YOLOv5+ms-testing+Classifier & 35.74 (\textcolor{green}{$\uparrow$ 0.84})& 57.31 (\textcolor{green}{$\uparrow$ 0.91}) \\
\hline
\end{tabular}
}
\end{center}
\caption{Ablation Study on VisDrone2021 testset-dev.}
\label{table:testset-dev}
\end{table}

\noindent \textbf{Effect of extra prediction head.}
Adding a detection head for tiny objects makes the number of layers of the original YOLOv5x change from 607 to 719, and GFLOPs from 219.0 to 259.0.
This of course increases the amount of calculation, but the mAP improvement is also very high. From Fig.~\ref{fig:detect} we can see that TPH-YOLOv5 performs well when detecting small objects, so the increasing in calculation is worthwhile.

\noindent \textbf{Effect of transformer encoder blocks.}
After using the transformer encoder block, the total layers of the model decrease from 719 to 705, and GFLOPs from 259.0 to 237.3. 
Use transformer encoder blocks can not only increase mAP, but also reduce the size of the network. At the same time, it also plays a role in the detection of dense objects and large objects.

\begin{figure}[htbp]
\begin{center}
\includegraphics[width=0.9\linewidth]{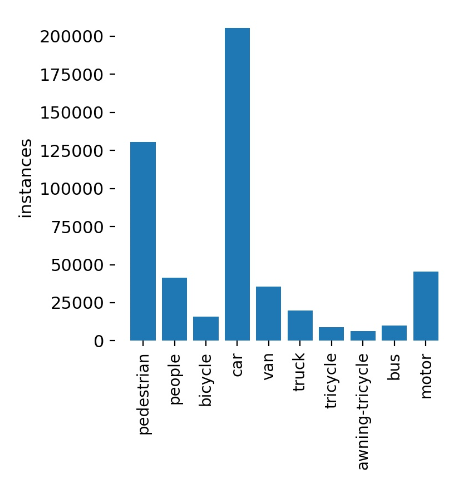}
\end{center}
   \caption{The number of labels of each category.}
\label{fig:num of labels}
\end{figure}

\begin{figure*}[htbp]
\begin{center}
\includegraphics[width=1\linewidth]{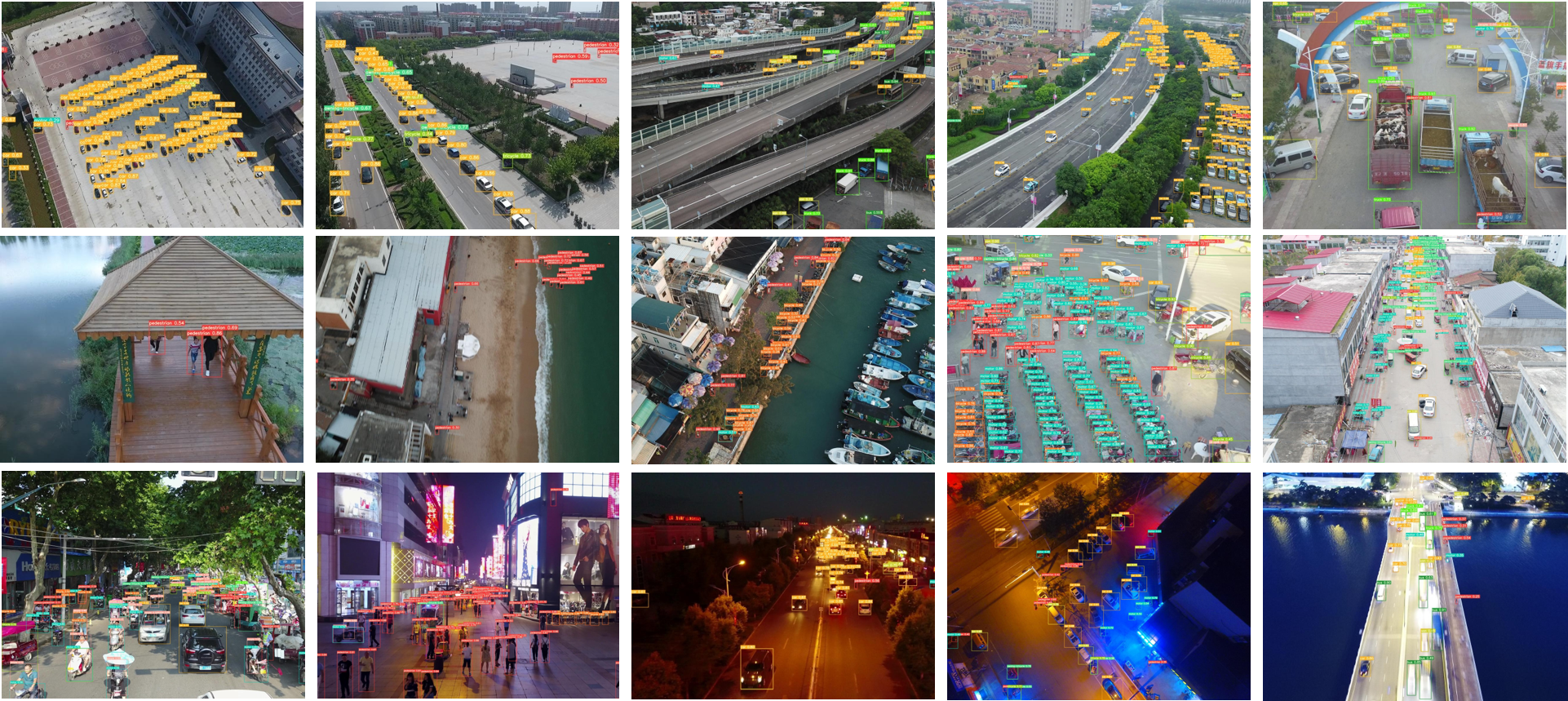}
\end{center}
   \caption{Some visualization results from our TPH-YOLOv5 on testset-challenge, different category use bounding boxes with different color. The performance is good at localization tiny objects, dense objects and objects blurred by motion. }
\label{fig:detect}
\end{figure*}

\noindent \textbf{Effect of model ensemble.}
We list the mAP of the final results of our five different models in each category and compared them with the fusion model in table~\ref{table:ensemble}. 
In training phrase, we use different input image sizes and change the weight of each category to make each model unique. So that the final ensemble model can get a relatively balanced result.
1) TPH-YOLOv5-1 use the input image size of 1920 and all categories have equal weights. 
2) TPH-YOLOv5-2 use the input image size of 1536 and all categories have equal weights. 
3) TPH-YOLOv5-3 use the input image size of 1920 and the weight of each category is related to the number of labels, which is shown in Fig.~\ref{fig:num of labels}. The more labels of a certain category, the lower the weight it is given.
4) TPH-YOLOv5-4 use the input image size of 1536 and the weight of each category is related to the number of labels. 
5) TPH-YOLOv5-5 use the backbone of YOLOv5l and use the input image size of 1536.

\begin{table*}[htbp]
\begin{center}
\resizebox{0.9\textwidth}{!}{ 
\begin{tabular}{|l|c|c|c|c|c|c|c|c|c|c|c|}
\hline
Methods & all & pedestrian & people & bicycle & car & van & trunk & tricycle & awning-tricycle & bus & motor\\
\hline\hline
TPH-YOLOv5-1          & 34.90 & 27.52 & 15.32 & 15.21 & 65.99 & 44.23 & 47.56 & 23.96 & 22.11 & 58.85 & 28.44 \\
TPH-YOLOv5-2          & 34.29 & 27.97 & 14.88 & 14.17 & 67.63 & 45.01 & 44.76 & 25.12 & 20.48 & 55.72 & 27.74  \\
TPH-YOLOv5-3          & 34.68 & 22.88 & 16.01 & 19.26 & 48.88 & 42.98 & 47.82 & 32.86 & 35.65 & 54.16 & 28.25 \\
TPH-YOLOv5-4          & 34.17 & 23.48 & 15.79 & 17.62 & 49.99 & 42.76 & 47.13 & 31.66 & 32.21 & 54.19 & 27.37 \\
TPH-YOLOv5-5          & 33.04 & 25.98 & 14.90 & 13.10 & 63.05 & 43.45 & 42.56 & 25.20 & 21.06 & 53.65 & 27.10 \\
TPH-YOLOv5 ensemble   & 37.32 & 29.00 & 16.75 & 15.69 & 68.94 & 49.79 & 45.16 & 27.33 & 24.72 & 61.80 & 30.90 \\
\hline
\end{tabular}
}
\end{center}
\caption{Comparison of TPH-YOLOv5 models‘ performances on VisDrone2021 testset-dev for each category.}
\label{table:ensemble}
\end{table*}

\noindent \textbf{Some detection result on VisDrone2021 testset-challenge.}
We have selected some representative images as the display of the test results. Fig.~\ref{fig:detect} shows the result of large objects, tiny objects, dense objects and the image covering a large area.

\section{Conclusion}
In this paper, we add some cutting-edge techniques \ie transformer encoder block, CBAM and some experienced tricks to YOLOv5 and form a state-of-the-art detector called TPH-YOLOv5, which is especially good at object detection in drone-captured scenarios. We refresh the record of VisDrone2021 dataset, our experiments showed that TPH-YOLOv5 achieved state-of-the-art performance in VisDrone2021 dataset. We have tried a large number of features, and used some of them to improve the accuracy of object detector. We hope this report can help developers and researchers get a better experience in the analysis and processing of drone-captured scenarios.


\section{Acknowledgments}
This work was supported by National Natural Science Foundation of China (62072021).

{\small
\bibliographystyle{ieee_fullname}
\bibliography{egbib}
}

\end{document}